\documentclass[5p,12pt,authoryear]{elsarticle}

\usepackage{amssymb}
\usepackage[acronym]{glossaries}
\usepackage{graphicx}

\journal{and accepted by Natural Language Processing}

\makeglossaries

\newacronym{ood}{OOD}{out-of-distribution}
\newacronym{nlp}{NLP}{Natural Language Processing}
\newacronym{ner}{NER}{Named Entity Recognition}
\newacronym{sa}{SA}{Sentiment Analysis}
\newacronym{bow}{BoW}{bag-of-words}
\newacronym{cbow}{CBoW}{continuous Bag-of-Words}
\newacronym{sltc}{SLTC}{Swedish Language Technology Conference}
\newacronym{ann}{ANN}{artificial neural network}
\newacronym{nn}{NN}{neural network}
\newacronym{lstm}{LSTM}{Long Short Term Memory Network}
\newacronym{sota}{SotA}{state-of-the-art}
\newacronym{nlg}{NLG}{Natural Language Generation}
\newacronym{mwe}{MWE}{Multi-Word Expression}
\newacronym{cnn}{CNN}{Convolutional Neural Network}
\newacronym{sw}{SW}{Simple Wiki}
\newacronym{mt}{MT}{Machine Translation}
\newacronym{gdc}{GDC}{Gothenburg Dialog Corpus}
\newacronym{t5}{T5}{Text-to-Text-Transfer Transformer}
\newacronym{roberta}{RoBERTa}{Robustly optimized BERT approach}
\newacronym{bert}{BERT}{Bidirectional Encoder Representations from Transformers}
\newacronym{mcc}{MCC}{Matthews Correlation Coefficient}
\newacronym{ai}{AI}{artificial intelligence}
\newacronym{xai}{XAI}{explainable artificial intelligence}
\newacronym{lime}{LIME}{Local Interpretable Model-agnostic Explanations}
\newacronym{bilstm}{Bi-LSTM}{Bi- Directional Long Short Term Memory Network}
\newacronym{rnn}{RNN}{Recurrent Neural Network}
\newacronym{ml}{ML}{machine learning}
\newacronym{hasoc}{HASOC}{hate speech and offensive content}
\newacronym{olid}{OLID}{offensive language identification dataset}
\newacronym{ig}{IG}{Integrated Gradient}
\newacronym{shap}{SHAP}{SHapley Additive exPlanations}
\newacronym{hs}{HS}{hate speech}
\newacronym{trac}{TRAC}{Trolling, Aggression and Cyberbullying}
\newacronym{hso}{HSO}{hate speech and offensive}
\newacronym{ooc}{OOC}{out-of-class}
\newacronym{os}{OS}{operating system}
\newacronym{lr}{LR}{learning rate}
\newacronym{multiwoz}{MultiWOZ}{Multi-Domain Wizard-of-Oz}
\newacronym{dialogpt}{DialoGPT}{Dialogue Generative Pre-trained Transformer}
\newacronym{bold}{BOLD}{Bias in Open-Ended Language Generation Dataset}
\newacronym{cuad}{CUAD}{Contract Understanding Atticus Dataset}
\newacronym{ledgar}{LEDGAR}{Labeled Electronic Data Gathering, Analysis, and Retrieval system}
\newacronym{genbit}{GenBiT}{
Gender bias in text toolkit}
\newacronym{imdb}{IMDB}{
Internet Movie Database}
\newacronym{mdgender}{MDGender}{
Multi-Dimensional Gender}
\newacronym{sbic}{SBICv2}{Social Bias Inference Corpus v2}
\newacronym{tp}{tp}{true positives}
\newacronym{tn}{tn}{true negatives}
\newacronym{fp}{fp}{false positives}
\newacronym{fn}{fn}{false negatives}
\newacronym{bat}{BAT}{Bias-Aware Thresholding}
\newacronym{tf}{TF}{term frequency}
\newacronym{mab}{MAB}{multi-axes bias dataset}
\newacronym{squad}{SQuADv2}{Stanford Question Answering Dataset}
\newacronym{copa}{COPA}{Choice Of Plausible Alternatives}
\newacronym{oodo}{OOD}{out-of-domain}
\newacronym{pos}{PoS}{part-of-speech}
\newacronym{pii}{PII}{personal identifiable information}
\newacronym{llm}{LLM}{large language model}

\begin{document}

\begin{frontmatter}



\title{Bipol: A Novel Multi-Axes Bias Evaluation Metric with Explainability for \acrshort{nlp}}


\author{Lama Alkhaled\textsuperscript{*}, Tosin Adewumi\textsuperscript{*$\dag$}, and Sana Sabah Sabry\\
\footnotesize*Joint first authors, \textsuperscript{$\dag$}corresponding author}
\affiliation{organization={ML Group, EISLAB, Luleå University of Technology},
            country={Sweden}\\
firstname.lastname@ltu.se}

\begin{abstract}
We introduce bipol, a new metric with explainability, for estimating social bias in text data.
Harmful bias is prevalent in many online sources of data that are used for training \acrfull{ml} models.
In a step to address this challenge we create a novel metric that involves a two-step process: corpus-level evaluation based on model classification and sentence-level evaluation based on (sensitive) \acrfull{tf}.
After creating new models to classify bias using \acrshort{sota} architectures, we evaluate two popular \acrshort{nlp} datasets (\acrshort{copa} and \acrshort{squad}) and the WinoBias dataset.
As additional contribution, we created a large English dataset (with almost 2 million labeled samples) for training models in bias classification and make it publicly available.
We also make public our codes.

\end{abstract}


\begin{highlights}
\item Introduction of bipol, the novel multi-axes bias estimation metric.
\item Public release of a new English large, labeled multi-axes bias dataset.
\item Release of multi-axes bias lexica, based on public sources.
\end{highlights}


\end{frontmatter}


\section{Introduction}
\label{intro}

Bias can be a difficult subject to tackle, especially as there are different opinions as to the scope of its definition \citep{Dhamala2021,hammersley1997bias}.
The origin of the word means a \textit{slant} or \textit{slope}.\footnote{etymonline.com/word/bias}
In this work, we define social bias as the unbalanced disposition (or prejudice) in favor of or against a thing, person or group, relative to another, in a way that is deemed as unfair \citep{adewumi2019conversational,antoniak2021bad,maddox2004perspectives}.\footnote{https://libguides.uwgb.edu/bias}
This is harmful bias and it is related to fairness.
In some quarters, bias also involves overgeneralization \citep{brigham1971ethnic,rudinger-etal-2018-gender,nadeem-etal-2021-stereoset}, fulfilling characteristic  \textit{2} of bias in the next paragraph.
Furthermore, some recent works have produced benchmark datasets with pairs of contrastive sentences (e.g. WinoBias \citep{zhao-etal-2018-gender}), which have been found to have a number of shortcomings that threaten their validity as measurement models for bias or stereotyping, as they often have ambiguities and unstated assumptions \citep{blodgett-etal-2021-stereotyping}.


As a motivation, we address the challenge of how bias in text data can be estimated along some of the many axes (or dimensions) of bias (e.g. race and gender).
Social bias in text usually has some of the following characteristics.\textsuperscript{2}

\begin{enumerate}
    \item It is heavily one-sided \citep{zhao-etal-2018-gender}, as will be observed with the results in this work.
    \item It uses extreme or inappropriate language \citep{rudinger-etal-2018-gender}.
    This forms the basis of the assumption (for some of the samples) in the two datasets used to create the new \acrfull{mab}, as discussed in Section \ref{datasets}.
    \item It is based on unsupported or unsubstantiated claims, such as stereotypes \citep{brigham1971ethnic}.
    \item It is entertainment-based or a form of parody or satire \citep{eliot2002personal}.
   
\end{enumerate}
 
\acrshort{ml} models pick these biases from the data they are trained on.
Although classification accuracy has been observed to fall with attempts at mitigating biases in data \citep{9174293,oneto2019taking,NIPS2017_b8b9c74a,speicher2018unified}, it is important to estimate and mitigate them, nonetheless.
This is because of the ethical implications and harm that may be involved for the disadvantaged group \citep{klare2012face,10.1145/3375627.3375820}.

\paragraph{Our contributions} We introduce a novel multi-axes bias estimation metric called \textit{bipol}.
The name \textit{\textbf{bipol}} emerged from the authors' combination of two words: \textit{\textbf{bi}as} and \textit{\textbf{pol}arity}.
Compared to other bias metrics, this is not limited in the number of bias axes it can evaluate and has explainability built in.
It will provide researchers with deeper insight into how to mitigate bias in data.
Our second contribution is the introduction of the new English \acrshort{mab} dataset.
It is a large, labeled dataset that is aggregrated from two other sources.
A third contribution is the multi-axes bias lexica we collected from public sources.
We perform experiments using \acrfull{sota} models to benchmark on the dataset.
Furthermore, we use the trained models to evaluate two common \acrshort{nlp} datasets (\acrshort{squad} \citep{rajpurkar-etal-2018-know} and (\acrshort{copa} \citep{roemmele2011choice}) and the WinoBias dataset \citep{zhao-etal-2018-gender}.
We make our models, codes, dataset, and lexica publicly available.\footnote{github.com/LTU-Machine-Learning/bipol}

The rest of this paper is structured as follows:
Section \ref{bipol} describes in detail the characteristics of the new metric.
Section \ref{datasets} gives details of the new \acrshort{mab} dataset.
Section \ref{experiments} explains the experimental setup.
Section \ref{results} presents the results and error analyses.
Section \ref{related} discusses some previous related work.
In Section \ref{conclusion}, we give concluding remarks.

\section{Bipol}
\label{bipol}
\textit{Bipol}, represented by Equation \ref{eq:eq1}, involves a two-step mechanism: the corpus-level evaluation (Equation \ref{eq:eq2}) and the sentence-level evaluation (Equation \ref{eq:eq3}).
It is a score between 0.0 (zero or undetected bias) and 1.0 (extreme bias).
This is further described below.

\begin{subequations}
\small

\begin{equation}
\mathit{b}=\begin{cases}
\mathit{b_{c}} . \mathit{b_{s}}, & \text{if $b_{s}>0$}\\
\mathit{b_{c}}, & \text{otherwise}
\end{cases}
\label{eq:eq1}
\end{equation}


\begin{equation}
\mathit{b_{c}} = 
\frac{tp + fp}{tp+fp+tn+fn}
\label{eq:eq2}
\end{equation}

\begin{equation}
\mathit{b_{s}} = \frac{1}{r} \sum_{t=1}^{r} 
{\left( \frac{1}{q} \sum_{x=1}^{q} {\left(
\frac{|\sum_{s=1}^{n} a_{s} - \sum_{s=1}^{m} c_{s}|}{\sum_{s=1}^{p} d_{s}}
\right)}_{x}
\right)}_{t}
\label{eq:eq3}
\end{equation}
\end{subequations}

\begin{enumerate}
    \item In step 1, a bias-trained model is used to classify all the samples for being biased or unbiased.
    The ratio of the biased samples (i.e. predicted positives) to the total samples predicted makes up this evaluation.
    When the true labels are available, this step is represented by Equation \ref{eq:eq2}.
    The predicted positives is the sum of the \acrfull{tp} and \acrfull{fp}.
    The total samples predicted is the sum of the \acrfull{tp}, \acrfull{fp}, \acrfull{tn}, and \acrfull{fn}.
    
    A more accurate case of the equation will be to have only the \acrshort{tp} evaluated (in the numerator), however, since we want comparable results to when \textit{bipol} is used in the "wild" with any dataset, we choose the stated version in \ref{eq:eq2} and report the positive error rate.
    Hence, in an ideal case, an \acrshort{fp} of zero is preferred.
    However, there's hardly a perfect classifier.
    It is also preferable to maximize \acrshort{tp} to capture all the biased samples, if possible.
    False positives exist in similar classification systems (such as hate speech detection, spam detection, etc) but they are still used \citep{adewumi2022t5,feng2018multistage,heron2009technologies,markines2009social}.
    New classifiers may also be trained for this purpose without using ours, as long as the dataset used is large and representative enough to capture the many axes of biases, as much as possible.
    Hence, \textit{bipol}'s two-step mechanism may be seen as a framework.
    
    \item In step 2, if a sample is positive for bias, it is evaluated token-wise along all possible bias axes, using all the lexica of sensitive terms.
    Table \ref{table:lexi} provides the lexica sizes.
    The lexica are adapted from public sources\footnote{merriam-webster.com/thesaurus/female,merriam-webster.com/thesaurus/male, en.wikipedia.org/wiki/List\_of\_ethnic\_slurs, en.wikipedia.org/wiki/List\_of\_religious\_slurs} and may be expanded as the need arises, given that bias terms and attitudes are ever evolving \citep{antoniak2021bad,haemmerlie1991goldberg}.
    They include terms that may be stereotypically associated with certain groups \citep{zhao-etal-2017-men,zhao-etal-2018-gender} and names associated with specific gender \citep{nangia-etal-2020-crows}.
    
    Examples of racial terms stereotypically associated with the white race (which may be nationality-specific) include \textit{charlie} (i.e. \textit{the oppressor}) and \textit{bule} (i.e. \textit{albino} in Indonesian) while \textit{darkey} and \textit{bootlip} are examples associated with the black race. 
    Additional examples from the lexica are provided in the appendix.
    Each lexicon is a text file with the following naming convention: \textit{axes\_type.txt}, e.g. \textit{race\_white.txt}.
    In more detail, step 2 (given by Equation \ref{eq:eq3}) involves finding the absolute difference between the two maximum summed frequencies (as lower frequencies cancel out) in the types of an axis (\(|\sum_{s=1}^{n} a_{s} - \sum_{s=1}^{m} c_{s}| \)).
    This is divided by the summed frequencies of all the terms ($d_{s}$) in that axis (\( \sum_{s=1}^{p} d_{s} \)).
    This operation is then carried out for all axes ($q$) and the average obtained (\( \frac{1}{q} \sum_{x=1}^{q} \)).
    Then it is carried out for all the biased samples ($r$) and the average obtained (\( \frac{1}{r} \sum_{t=1}^{r} \) ).


\end{enumerate}

\begin{table}[h]
\centering
\resizebox{\columnwidth}{!}{%
\begin{tabular}{lccc}
\hline
\textbf{Axis} & \textbf{Axis type 1} & \textbf{Axis type 2} & \textbf{Axis type 3}\\
\hline
Gender & 76 (female) & 46 (male) & \\
Racial & 84 (black) & 127 (white) & \\
Religious & 180 (christian) & 465 (muslim) & 179 (hindu)\\
\hline
\end{tabular}
}
\caption{\label{table:lexi}
Lexica sizes. These may be expanded.
}
\end{table}


The use of the two-step process minimizes the possibility of wrongly calculating the metric on a span of text solely because it contains sensitive features.
For example, given the sentences below\footnote{These are mere examples. People's names have been anonymized with the PERSON entity in the dataset}

\begin{quote}
\small
    1. \textit{A nurse should wear her mask as a pre-requisite.}\\
    2. \textit{Veronica, a nurse, wears her mask as a pre-requisite.}\\
    3. \textit{This nurse wears her mask as a pre-requisite.}\\
    4. \textit{A nurse should wear his or her mask as a pre-requisite.}\\
\end{quote}
the first one should be classified as biased by a model in the first step, ideally, because the sentence assumes a nurse should be female.
The second step can then estimate the level of bias in that sentence, based on the lexica.
In the second example, a good classifier should not classify this as biased since the coreference of \textit{Veronica} and \textit{her} are established, with the assumption that \textit{Veronica} identifies as a female name.
Similarly, the third example should not be classified as biased by a good classifier because \textit{"This"} refers to a specific nurse and more context or extreme language may be required to classify it otherwise, as discussed in the introduction.
Note that the second example becomes difficult to classify, even for humans, if \textit{Veronica} was anonymized, say with a \acrfull{pos} tag.
In the case of the fourth example, an advantage of \textit{bipol} is that even if it is misclassifed as biased, the sentence-level evaluation will evaluate to zero because the difference between the maximum frequencies of the two types (\textit{his} and \textit{her}) is \textit{1 - 1 = 0}.
\textit{Bipol} does not differentiate explicitly whether the bias is in favour of or against a targeted group.

\paragraph{Strengths of \textit{bipol}}

\begin{enumerate}
    \item It is relatively simple to calculate.
    \item It is based on existing tools (classifiers and lexica), so it is straight-forward to implement. 
    \item It is a two-step process that captures both semantic and \acrfull{tf} aspects of text.
    \item It is flexible, as it has no limits in the number of axes or \acrshort{tf} that can be included.
    \item Its explainability makes up for what is not obvious from a single score.
    The goal of explainable \acrshort{ai} is to enable end users to understand \acrshort{ai} decisions and there exist many forms of explainability, including explanation by example, textual explanation, and visualisation of the decision space, among others \citep{gunning2021darpa}.
    
    \textit{Bipol}, hence, provides a dictionary of lists of the sensistive term frequencies in any evaluated data, providing the opportunity to visualize them in plots.
    For example, the magnitude of the difference between term frequencies in an axis is not immediately obvious from a single score of, say, 1. This is because \( (1-0)/1 = (1,000-0)/1,000 = 1 \).
    As an example, if \textit{he} has a frequency of 1 while \textit{she} has 0 in one instance, it is the same score  of 1 if they have 1,000 and 0, respectively, in another instance.

\end{enumerate}

\paragraph{Weakness of \textit{bipol}}

\begin{enumerate}
    \item Although one of its strengths is that it is based on existing tools, this happens to also be a weakness, since the limitations of these tools also limit its accuracy.
    
\end{enumerate}

\section{Datasets}
\label{datasets}
\subsection*{The new \acrshort{mab} dataset}
This English bias-detection dataset has a total size of 1,946,975 samples, as given in Table \ref{table:genbias}.
It is a combination of the Jigsaw\footnote{medium.com/jigsaw/creating-labeled-datasets-and-exploring-the-role-of-human-raters-56367b6db298} (of 1,902,194 samples) and the \acrfull{sbic} (of 147,139 samples) by \cite{sap-etal-2020-social}.
It has 12 explicit bias axes (from the combination of 7 axes from the Jigsaw and 5 additional axes from \acrshort{sbic}).
This makes it one of the largest annotated datasets for bias detection, especially when compared to \acrfull{bold} with 23,679 samples \citep{Dhamala2021} or HolisticBias with 459,758 samples \citep{smith-etal-2022-im}.
The large size of the dataset increases the chances of training a classifier to identify a broad range of biased cases.

In creating the data, we dropped duplicates and anonymized \acrfull{pii} of 1,373,088 instances of people's names with \textit{PERSON}.
Examples in the \acrshort{mab} are given in Table \ref{table:mabsamples}.

\begin{table}[h]
\small
\centering
\begin{tabular}{lccc}
\hline
\textbf{Set} & \textbf{Biased} & \textbf{Unbiased} & \textbf{Total}\\
\hline
Training & 533,544 & 1,209,433 & 1,742,977 \\
Validation & 32,338 & 69,649 & 101,987\\
Test & 33,470 & 68,541 & 102,011\\
\hline
 & 599,352 & 1,347,623 & 1,946,975 \\
 \hline
\end{tabular}
\caption{\label{table:genbias}
\acrshort{mab} dataset split
}
\end{table}

\begin{table}[h]
\tiny
\centering
\begin{tabular}{p{0.48\linewidth} | p{0.12\linewidth} | 
 p{0.1\linewidth} | p{0.12\linewidth}}
\hline
\textbf{comment\_text} & \textbf{label} & \textbf{old\_id} & \textbf{id}  
\\
\hline
This b*tch is nuts. Who would read a book by a woman. & biased & 239612 & 106351 \\
What do you call lingerie for Arabs? Socks. & biased & none & 1355035 \\
I am assuming a sequel is in the works, then?" & unbiased & 282386 & 613423\\
What are the chances that a Jew had the number 8675309? 1 in 6 million & biased & none & 812633\\

\hline
\end{tabular}
\caption{\label{table:mabsamples}
Examples from the \acrshort{mab}
}
\end{table}

The Jigsaw is a multipurpose dataset that assumes that bias correlates with toxicity.
This assumption is realistic and has been used in previous work in the literature \citep{nangia-etal-2020-crows}.
Hence, we automatically annotated as \textit{biased} the \textit{target} and \textit{toxicity} columns in the training and test sets, respectively, with values greater than or equal to the bias threshold of 0.1 (on a scale from 0 to 1) while those below are automatically annotated as \textit{unbiased}.
The rationale for choosing the threshold of 0.1 is that this threshold represents about 494 human annotators out of the maximum 4,936 in some instances and it seems inappropriate to dismiss their view.
It is also based on our random inspection of several examples in the dataset.
For example, the comment below, which we consider biased, has a \textit{target} greater than 0.1 and much lesser than 0.5.

\begin{quote}
\small
    \textit{In 3 years from now, the Alaska Permanent Fund Dividend will be ZERO \$\$\$. Democrats will moan, wail, and scream that there is no more OTHER PEOPLES' MONEY to FREE GIFT.  Alaskans will have to go back to living on what money they earn for themselves. The oil boom is over. It's bust time in Alaska.}
\end{quote}
In addition, adopting a threshold higher than 0.1 will result in further imbalance in the dataset in favour of unbiased samples.

The \acrshort{sbic} dataset follows a similar assumption as the Jigsaw.
We use the aggregrated version of the dataset and the same bias threshold for the \textit{offensiveYN} column in the sets.
In the Jigsaw, we retained the old IDs so that we can always trace back useful features to the original data source, but the \acrshort{sbic} did not use IDs.
The \acrshort{mab} data statement is provided in the appendix (\ref{datacard}).
More details of the two base datasets are given in the following paragraphs.

\paragraph{Jigsaw}
The Jigsaw dataset came about as a result of annotations by the civil comments platform.
It has the following axes: gender, sexual orientation, religion, race/ethnicity, disability, and mental illness.
The average scores given by all annotators is calculated to get the final values for all the labels.
It contains 1,804,874 comments in the training set and 97,320 comments in the test set.
A small ratio (0.0539) was taken from the training set as part of the validation set for the \acrshort{mab} because the Jigsaw has no validation set and we wanted a validation set that is representative of the test set in size.
The Jigsaw was annotated by a total of almost 9,000 human raters, with a range of three to ten raters on average per comment.
It is under CC0 licence in the public domain.

\paragraph{\acrshort{sbic}} The dataset covers a variety of social biases implied in text, along the following axes: gender/sexuality, race/ethnicity, religion/culture, social/political, disability body/age, and victims.
Each split of the dataset has an aggregated-per-post version.
The annotations in \acrshort{sbic} showed 82.4\% pairwise agreement and Krippendorf $\alpha$=0.45 on average.
There are no usernames in the dataset.
The \acrshort{sbic} is licensed under the CC-BY 4.0 license.
The data is drawn from online posts from the following sources:
\begin{itemize}
    \item r/darkJokes, r/meanJokes, r/offensiveJokes (r: reddit)

    \item Reddit microaggressions \citep{breitfeller-etal-2019-finding}

    \item Toxic language detection Twitter corpora \citep{waseem-hovy-2016-hateful,davidson2017automated,founta2018large}

    \item Data scraped from hate sites (Gab, Stormfront)

\end{itemize}

\section{Materials and Methods}
\label{experiments}
All the experiments were conducted on two shared Nvidia DGX-1 machines running Ubuntu 18 and 20 with 8 × 32GB V100 and 8 × 40GB A100 GPUs, respectively.
Each evaluation experiment is conducted twice and the average results reported.
Wandb \citep{wandb}, the experiment tracking tool, runs for 16 counts with bayesian optimization to suggest the best hyper-parameter combination for the initial learning rate (1e-3 - 2e-5) and epochs (6 - 10), given the importance of hyper-parameters \citep{adewumi2020word2vec}.
These are then used to train the final models (on the Jigsaw, \acrshort{sbic} and \acrshort{mab}), which are then used to evaluate their test sets, the test set of WinoBias, the \textit{context} of the \acrshort{squad} validation set, and the \textit{premise} of the \acrshort{copa} training set (since models learn from training set).
The F1 metric that we report is given by:
\( \frac{2*\acrshort{tp}}{2*\acrshort{tp}+\acrshort{fp}+\acrshort{fn}} \)

\begin{figure*}[h]
\centering
\includegraphics[width=1\textwidth]{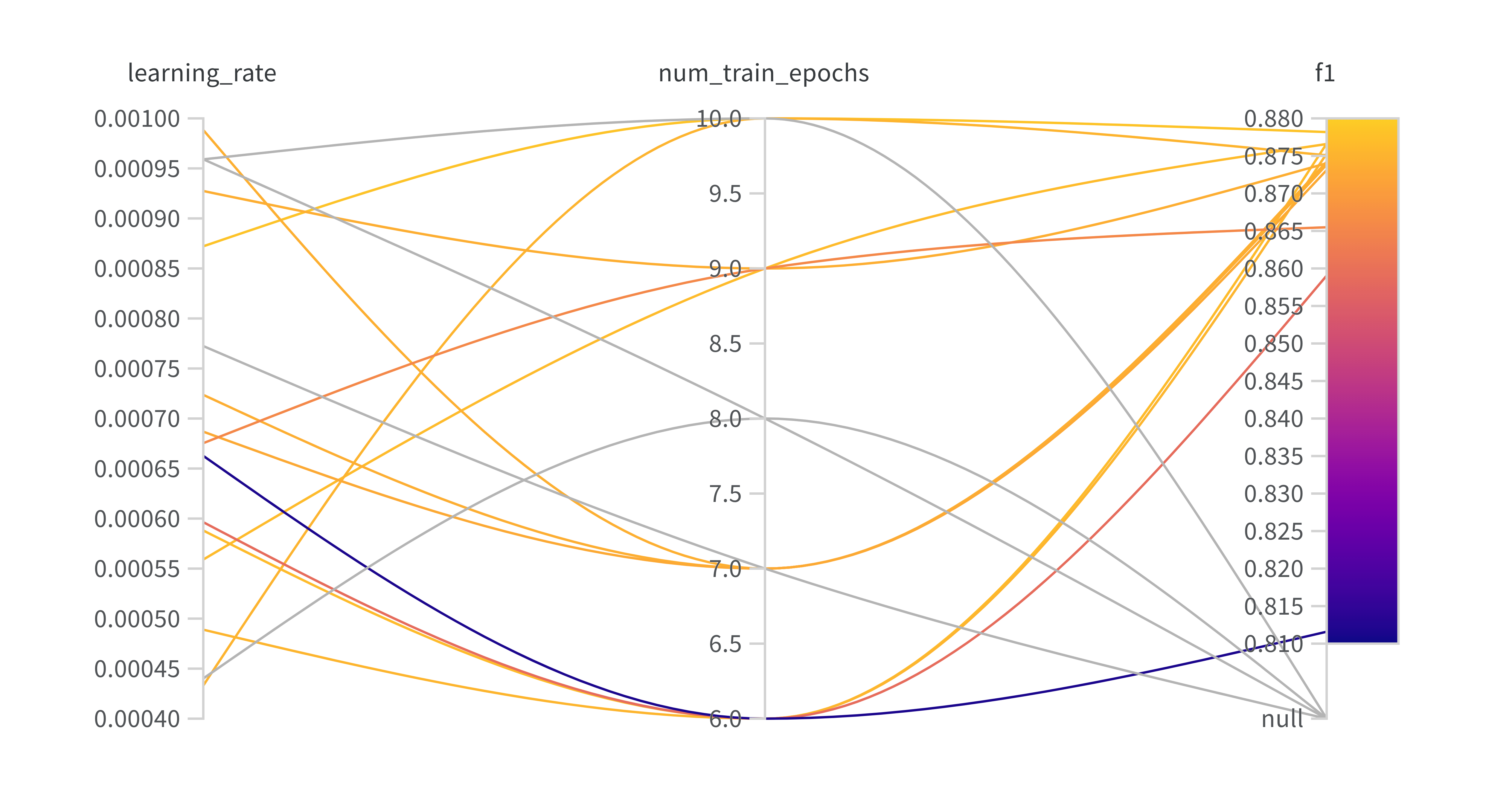}
\caption{Hyper-parameter optimization parallel coordinates for DeBERTa on \acrshort{mab}}
\label{fig:img1}
\end{figure*}

We use the pretrained base models of \acrshort{roberta} \citep{liu2019roberta}, DeBERTa \citep{he2021deberta} and Electra \citep{clark2020electra}, from the HuggingFace hub \citep{wolf-etal-2020-transformers}.
Figure \ref{fig:img1} shows the wandb exploration for DeBERTa on \acrshort{mab} in parallel coordinates.
Average training time ranges from 41 minutes to 3 days, depending on the data size.
Average test set evaluation time ranges from 4.8 minutes to over 72.3 hours.\footnote{when cpulimit is enforced, in fairness to other users.}

Although recent work has shown that WinoBias \citep{zhao-etal-2018-gender} contains grammatical issues, incorrect or ambiguous labels, and stereotype conflation, among other limitations \citep{blodgett-etal-2021-stereotyping}, we evaluate it using \textit{bipol}.
It is noteworthy that WinoBias is designed for coreference systems and accuracy is typically reported while \textit{bipol} is based on classifying text and analysing the sensitive term frequencies.


\section{Results and Discussion}
\label{results}

\begin{table*}[h!]
\tiny
\centering
\begin{tabular}{ccccccccc}
\hline
&  &  & \multicolumn{2}{c}{\textbf{macro F1} $\uparrow$ (s.d.)} & \multicolumn{3}{c}{\textbf{bipol level} $\downarrow$ (s.d.)} & \textbf{error rate} $\downarrow$ \\
\acrshort{roberta} & \textbf{axes} $\uparrow$ & \textbf{samples evaluated} & \textbf{dev} &  \textbf{test} & \textbf{corpus} & \textbf{sentence} & \textbf{bipol \textit{(b)}} & \textbf{\acrshort{fp}/(\acrshort{fp}+\acrshort{tp})} \\
\hline
 Jigsaw & 7 & 97,320 & 0.88 (0) & 0.778 (0) & 0.244 & 0.919 & 0.225 (0) & 0.236\\
\acrshort{sbic} & 11 & 4,691 & 0.763 (0.004) & 0.796 (0.004) & 0.755 & 0.711 & 0.538 (0.06) & 0.117\\
\acrshort{mab} & 12 & 102,011 & 0.877 (0) & 0.780 (0) & 0.246 & 0.925 & 0.227 (0) & 0.198 \\
\acrshort{copa} & - & 400 & &  & 0.03 & 0.917 & 0.027 (0) & $>$ 0.198\\
\acrshort{squad} & - & 1,204 & &  & 0.002 & 0 & 0.002 (0) & $>$ 0.198\\
WinoBias & 1 & 1,584 & & & 0.029 & 0.978 & 0.028 (0) & $>$ 0.198\\
 \\ \hline

DeBERTa & & & & & & & & \\
\hline

Jigsaw & 7 & 97,320 & 0.877 (0.004) & 0.771 (0) & 0.239 & 0.914 & 0.218 (0) & 0.222\\
\acrshort{sbic} & 11 & 4,691 & 0.767 (0) & 0.83 (0) & 0.754 & 0.712 & 0.537 (0) & 0.116\\
\acrshort{mab} & 12 & 102,011 & 0.876 (0.001) & 0.773 (0) & 0.239 & 0.923 & 0.22 (0) & 0.2\\
\acrshort{copa} & - & 400 & &  & 0.035 & 1 & 0.035 (0) & $>$ 0.2\\
\acrshort{squad} & - & 1,204 & &  & 0.007 & 0.883 & 0.006 (0) & $>$ 0.2\\
WinoBias & 1 & 1,584 & &  & 0.011 & 0.944 & 0.011 (0) & $>$ 0.2\\
 \\ \hline
 
 Electra & & & & & & & & \\
\hline
Jigsaw & 7 & 97,320 & 0.88 (0) & 0.769 (0) & 0.226 & 0.916 & 0.207 (0) & 0.216\\
\acrshort{sbic} & 11 & 4,691 & 0.712 (0.002) & 0.828 (0) & 0.706 & 0.667 & 0.471 (0) & 0,097\\
\acrshort{mab} & 12 & 102,011 & 0.875 (0) & 0.777 (0) & 0.241 & 0.925 & 0.223 (0) & 0.196\\
\acrshort{copa} & - & 400 & &  & 0.028 & 0.909 & 0.025 (0) & $>$ 0.196\\
\acrshort{squad} & - & 1,204 & &  & 0.004 & 0.587 & 0.002 (0) & $>$ 0.196\\
WinoBias & 1 & 1,584 & &  & 0.016 & 1 & 0.016 (0) & $>$ 0.196\\

\hline
\end{tabular}
\caption{\label{table:res}
\footnotesize Average F1 and \textit{bipol} scores. Lower is better for \textit{bipol} and the positive error rate.
\acrshort{copa}, \acrshort{squad}, and WinoBias are evaluated with the \acrshort{mab}-trained models and do not have F1 scores since we do not train on them.
}
\end{table*}

\begin{figure*}[h!]
\centering
\includegraphics[width=1\textwidth]{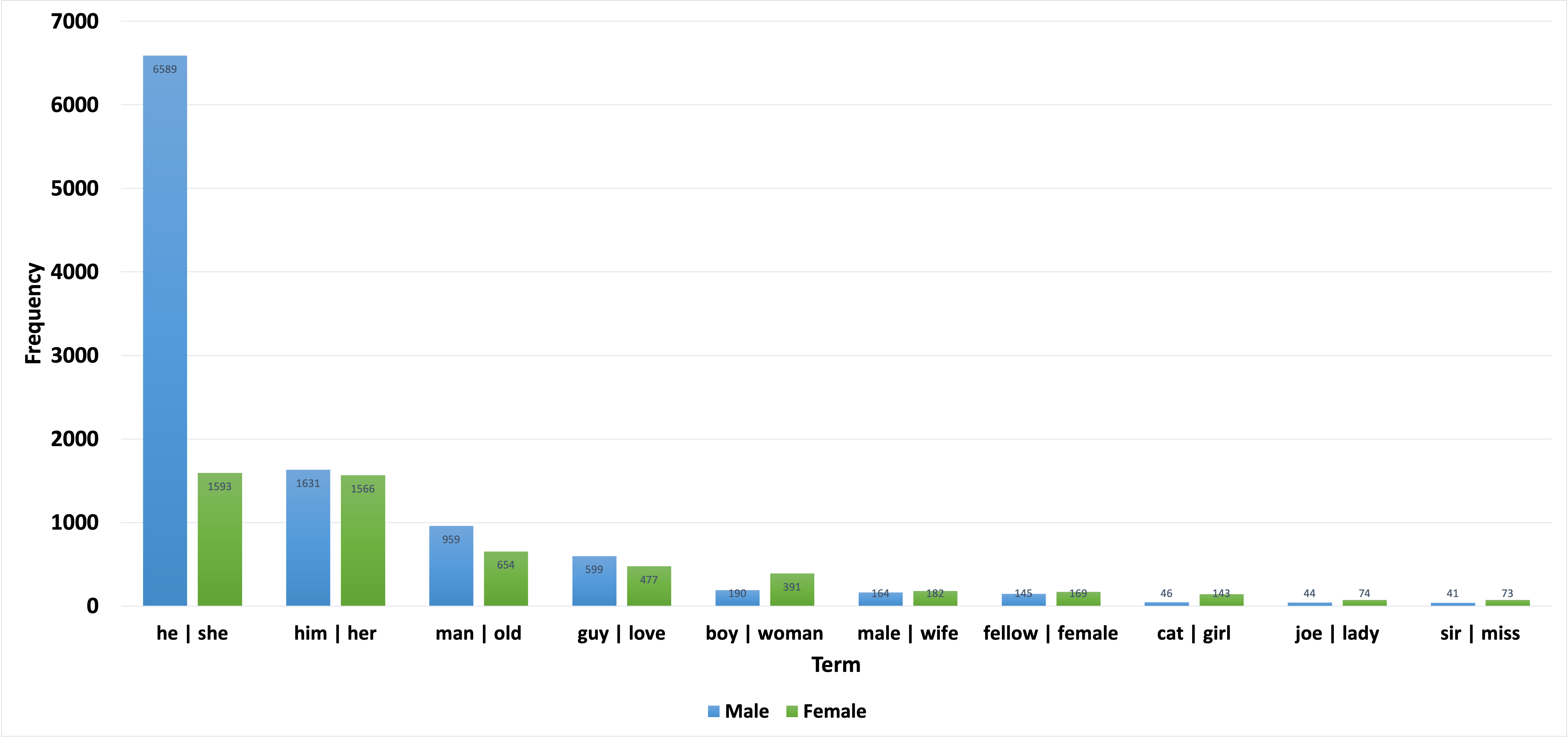}
\caption{\footnotesize Top-10 gender frequent terms influencing \textit{bipol} in the \acrshort{mab} test set after \acrshort{roberta} classification.
Terms like \textit{love \& old} are associated with the female gender according to the lexica.
However, when such subjective words are removed or put in both the male \& female lexica, they cancel out from influencing \textit{bipol}.}
\label{fig:top10rob}
\end{figure*}

\begin{figure*}[h!]
\centering
\includegraphics[width=1\textwidth]{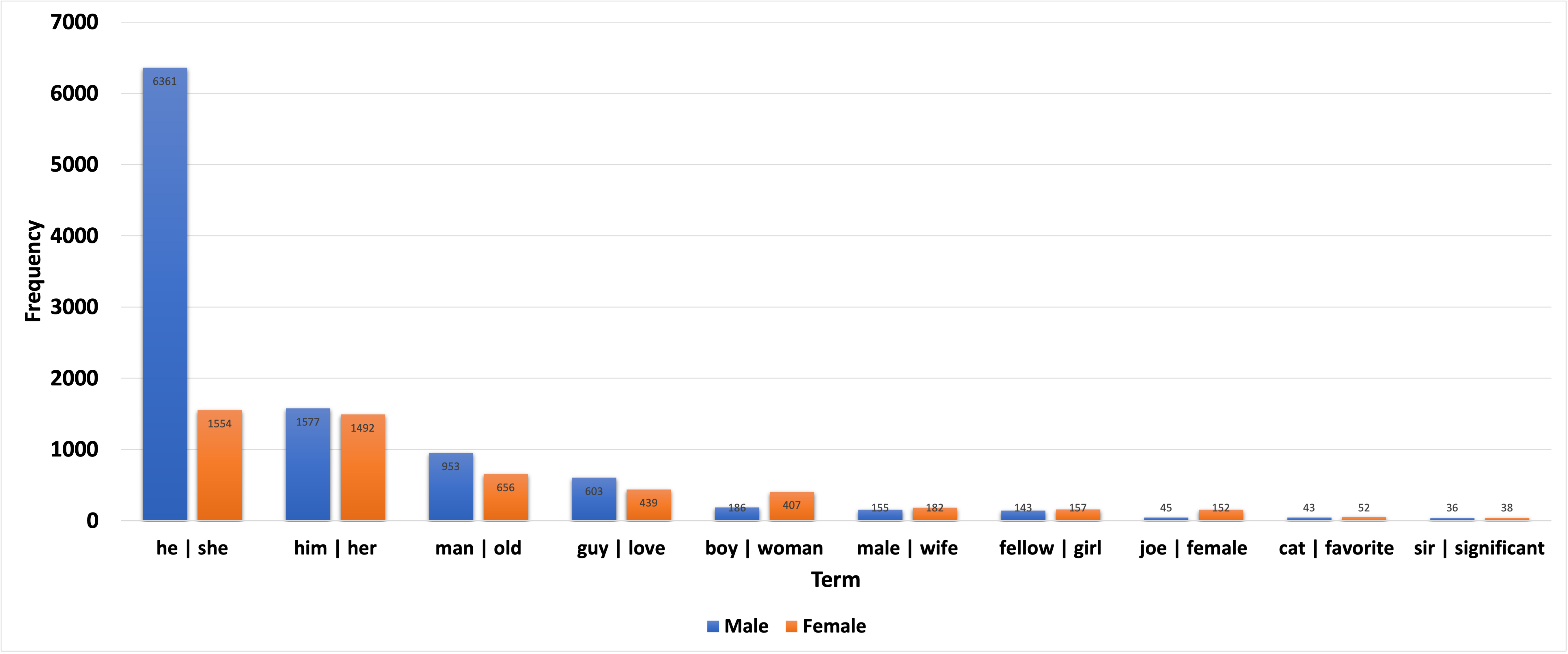}
\caption{\footnotesize Top-10 gender frequent terms influencing \textit{bipol} in the \acrshort{mab} test set after DeBERTa classification.
Terms like \textit{love \& old} are associated with the female gender according to the lexica.
However, when such subjective words are removed or put in both the male \& female lexica, they cancel out from influencing \textit{bipol}.}
\label{fig:top10}
\end{figure*}

\begin{figure*}[h!]
\centering
\includegraphics[width=1\textwidth]{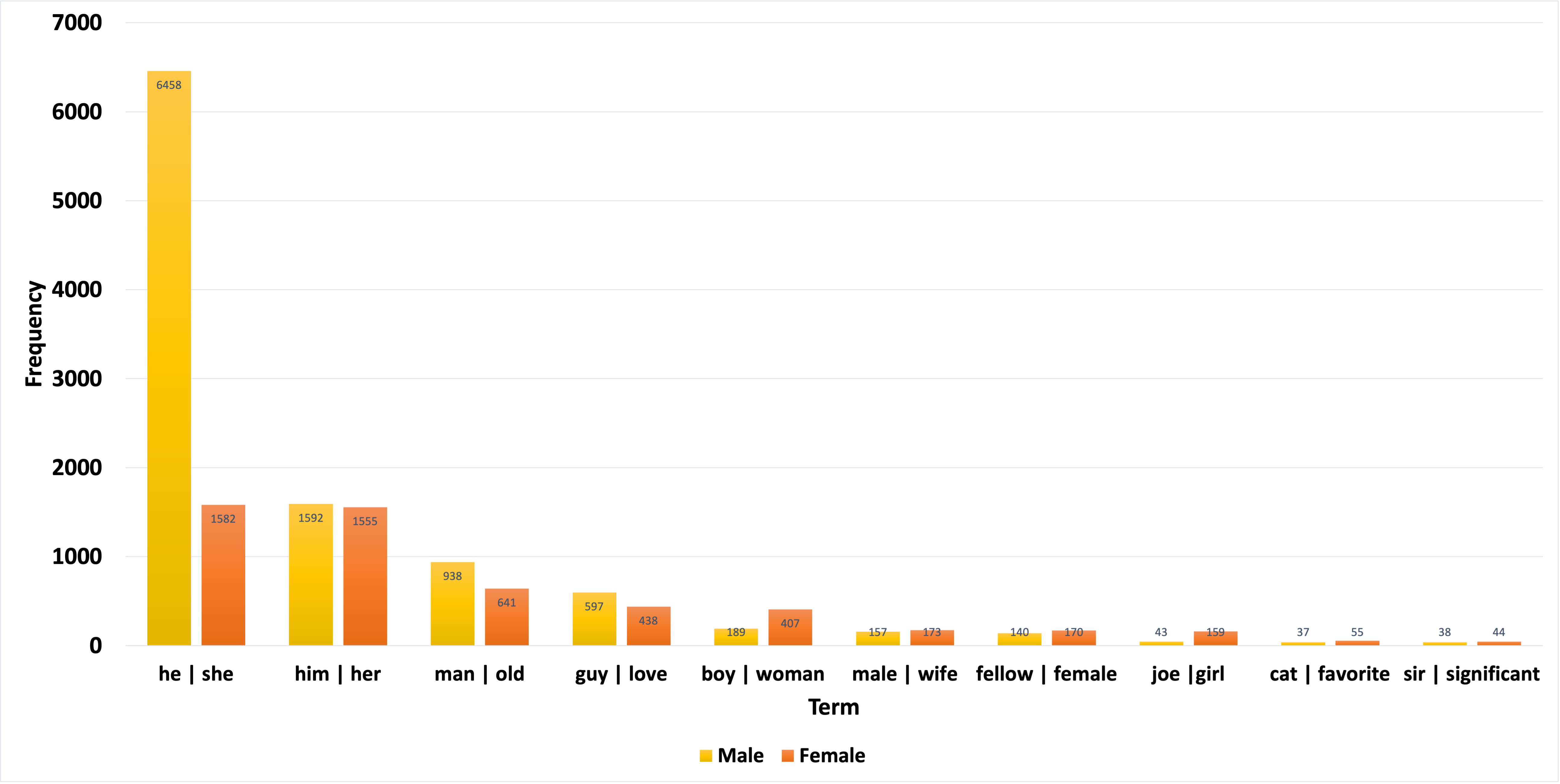}
\caption{\footnotesize Top-10 gender frequent terms influencing \textit{bipol} in the \acrshort{mab} test set after Electra classification.
Terms like \textit{love \& old} are associated with the female gender according to the lexica.
However, when such subjective words are removed or put in both the male \& female lexica, they cancel out from influencing \textit{bipol}.}
\label{fig:top10elec}
\end{figure*}

From Table \ref{table:res}, we observe that the results across the three models (\acrshort{roberta}, DeBERTa, \& Electra) for the datasets with training data (Jigsaw, \acrshort{sbic}, \& \acrshort{mab}) follow similar trends with regards to all the metrics.
They also show the 3 highest \textit{bipol} scores, as expected, since these datasets are designed to contain bias.
Indeed, all the datasets apparently contain bias but this is less common in \acrshort{copa}, WinoBias, and \acrshort{squad}.
As expected, \acrshort{mab} has slightly more bias compared to the Jigsaw, which is of comparable size, because additional bias was added from the \acrshort{sbic}.
\acrshort{sbic} shows more than 100\% \textit{bipol} increase over any of the other datasets - suggesting it contains much more bias relative to its size.
We also observe from the test set results that \acrshort{roberta} appears to be the best classifier except with \acrshort{sbic}, possibly because of the suggested hyper-parameters.
The two-sample t-test of the difference of means between the Jigsaw and \acrshort{mab}, across the 3 models, have \textit{p} values $<$ 0.0001, for 0.05 alpha, hence, the results are statistically significant.


To choose the preferred trained model for evaluating the other datasets (\acrshort{copa}, \acrshort{squad}, \& WinoBias), which therefore have no F1 scores, we prioritize the number of axes and data size, as this is more representative and likely to generalize better.
This is the reason why we used the \acrshort{mab}-trained models and provide the error rates, which indicate a lower bound of error for these datasets.

While there are a few metrics for estimating bias in text data, most are focused on gender bias, lexica and their libraries are unavailable, making such impossible to compare with \textit{bipol}. For example, GenBiT \citep{sengupta2021genbit} and Gender Gap Tracker \citep{asr2021gender} are lexica-based, which represent the non-semantic half of the method of \textit{bipol}, and are, therefore, less accurate than \textit{bipol}.

\subsection{\textit{Bipol} Explainability}
\textit{Bipol} generates a dictionary of lists of term frequencies, based on the lexica, that explains the score.
For example, a snapshot of the explainability dictionary of lists of terms, which produced the chart of top-10 gender frequent terms in Figure \ref{fig:top10}, is given in the following block.

\begin{quote}
\tiny
    \{'gender': [{' she ': 1554, ' her ': 1492, ' woman ': 407, ' lady ': 65, ' female ': 152, ' girl ': 157, ' skirt ': 5, ' madam ': 0, ' gentlewoman ': 0, ' madame ': 2, ' dame ': 3, ' gal ': 5, ' maiden ': 0, ' maid ': 2, ' damsel ': 0, ' senora ': 0, ' lass ': 0, ' beauty ': 16, ' ingenue ': 0, ' belle ': 0, ' doll ': 7, ' señora ': 0, ' senorita ': 0, ' lassie ': 0, ' ingénue ': 0, ' miss ': 67, ' mademoiselle ': 1, ' señorita ': 0, ' babe ': 3, ' girlfriend ': 32, ' lover ': 12, ' mistress ': 5, ' ladylove ': 0, ' inamorata ': 0, ' gill ': 1, ' old ': 656, ' beloved ': 16, ' dear ': 35, ' sweetheart ': 4, ' sweet ': 25, ' flame ': 5, ' love ': 439, ' valentine ': 1, ' favorite ': 52, ' moll ': 0, ' darling ': 8, ' honey ': 9, ' significant ': 38, ' wife ': 182, ' wifey ': 0, ' missus ': 0, ' helpmate ': 0, ' helpmeet ': 0, ' spouse ': 15, ' bride ': 1, ' partner ': 30, ' missis ': 0, ' widow ': 5, ' housewife ': 1, ' mrs ': 8, ' matron ': 0, ' soul ': 34, ' mate ': 5, ' housekeeper ': 1, ' dowager ': 0, ' companion ': 1, ' homemaker ': 0, ' consort ': 1, ' better half ': 1, ' hausfrau ': 0, ' stay-at-home ': 0}, {' he ': 6361, ' him ': 1577, ' boy ': 186, ' man ': 953, ' male ': 155, ' guy ': 603, ' masculine ': 4, ' virile ': 0, ' manly ': 4, ' man-sized ': 0, ' hypermasculine ': 0, ' macho ': 3, ' mannish ': 0, ' manlike ': 0, ' man-size ': 0, ' hairy-chested ': 0, ' butch ': 0, ' ultramasculine ': 0, ' boyish ': 0, ' tomboyish ': 0, ' hoydenish ': 0, ' amazonian ': 0, ' gentleman ': 13, ' dude ': 64, ' fellow ': 143, ' cat ': 43, ' gent ': 0, ' fella ': 2, ' lad ': 1, ' bloke ': 0, ' bastard ': 9, ' joe ': 45, ' chap ': 2, ' chappie ': 0, ' hombre ': 0, ' galoot ': 0, ' buck ': 25, ' joker ': 2, ' mister ': 3, ' jack ': 20, ' sir ': 36, ' master ': 26, ' buddy ': 25, ' buster ': 3}], 'racial': [{' nigga ': 61, ' negro ': 24, ... }\}

\end{quote}

From the bar charts (Figures \ref{fig:top10rob}, \ref{fig:top10} \& \ref{fig:top10elec}), we observe that the \acrshort{mab} dataset has a strong male bias.
In Figure \ref{fig:top10rob}, the top male term ('he') has a frequency of 6,589 while 'she' has only 1,593.
This follows a similar observation with other datasets \citep{fuertes2007corpus} or OneNote 5.0, a resource for training co-reference systems, that entities with gendered pronouns are over 80\% male \citep{zhao-etal-2018-gender}.
Furthermore, when highly subjective terms like \textit{love, old, favorite,} and \textit{significant} that are associated with the female gender in the lexica are removed or put in both the male and female lexica, they cancel out from influencing \textit{bipol}.
We note that WinoBias is limited to only gender, unlike \acrshort{squad}, which also reflects religious bias, as explained in their \textit{bipol} dictionaries of lists.
The artefacts of dictionaries of lists are publicly available.\textsuperscript{3} 
This shows the effectiveness of \textit{bipol} for capturing multiple axes.

\subsection{Qualitative Results}
Some qualitative examples of perceived correct predictions in COPA using the \acrshort{mab}-trained DeBERTa model are given in Table \ref{table:corrpred}.
We chose the DeBERTa-\acrshort{mab} model for this analysis because it has the lowest \textit{bipol} out of the three \acrshort{mab}-trained models since this presents the most challenging case to find successful examples. 
It appears the term \textit{cowgirl}, in the first example in the table, is perceived as gendered, the same way chairman or spokesman is perceived as such \citep{romaine2001corpus,fuertes2007corpus}.

\begin{table}[h]
\tiny
\centering
\begin{tabular}{p{0.58\linewidth} | p{0.12\linewidth} | p{0.1\linewidth}}
\hline
\textbf{examples} & \textbf{label} & \textbf{csv id}  
\\
\hline
the cowgirl threw her lasso towards the horse. & biased & 132 \\
\hline
the flashlight was dead. & unbiased & 136 \\
\hline
i wrote a thankyou note to my grandmother & unbiased & 139\\
\hline
\end{tabular}
\caption{\label{table:corrpred}
Examples of correct COPA predictions using \acrshort{mab}-trained DeBERTa model.
}
\end{table}

Some qualitative examples of correct predictions in \acrshort{squad} using the \acrshort{mab}-trained DeBERTa model are given in Table \ref{table:corrpred2}.
It seems the third example (id: 989) in the table has many different viewpoints that it's hard to find a justification for the model not to classify it as biased for at least one of them.

\begin{table}[h]
\tiny
\centering
\begin{tabular}{p{0.58\linewidth} | p{0.12\linewidth} | p{0.1\linewidth}}
\hline
\textbf{examples} & \textbf{label} & \textbf{csv id}  
\\
\hline
many locals and tourists frequent the southern california coast for its popular beaches and the desert city of palm springs is popular for its resort feel and nearby open spaces. & unbiased & 96 \\
\hline
rugby is also a growing sport in southern california particularly at the high school level with increasing numbers of schools adding rugby as an official school sport. & unbiased & 126\\
\hline
moderate and reformist islamists who accept and work within the democratic process include parties like the tunisian ennahda movement. jamaateislami of pakistan is basically a sociopolitical and democratic vanguard party but has also gained political influence through military coup détat in past. the islamist groups like hezbollah in lebanon and hamas in palestine participate in democratic and political process as well as armed attacks seeking to abolish the state of israel. radical islamist organizations like alqaeda and the egyptian islamic jihad and groups such as the taliban entirely reject democracy often declaring as kuffar those muslims who support it see takfirism as well as calling for violentoffensive jihad or urging and conducting attacks on a religious basis. & biased & 989 \\
\hline
\end{tabular}
\caption{\label{table:corrpred2}
Examples of correct \acrshort{squad} predictions using \acrshort{mab}-trained DeBERTa model.
}
\end{table}

\subsection{Error Analysis}
Table \ref{table:matrix} shows the prediction distribution for the models trained on \acrshort{mab}.
Unbiased samples are more easily detected in the dataset because there are more of these in the training set.
One way to improve the performance and the \acrshort{mab} dataset is to upsample the biased class.
This may be done through counter-factual data augmentation (CDA) or sentence completion through generative models.
Although \textit{bipol} is designed to be data-agnostic, it is important to note that estimating bias on \acrfull{oodo} datasets may result in less stellar performances.
This is because the trained \acrshort{mab} models are based on \acrshort{mab}'s 12 explicit bias axes.
Some qualitative examples of perceived incorrect predictions in \acrshort{copa} using the \acrshort{mab}-trained DeBERTa model are given in Table \ref{table:incorrpred}.
The second example (id: 71), particularly, may be considered incorrect since the definite article "the" is used to identify the particular subject "terrorist".

\begin{table}[h]
\small
\centering
\begin{tabular}{lcccc}
\hline
\textbf{Model-Data} & \textbf{\acrshort{tn}} & \textbf{\acrshort{fp}} & \textbf{\acrshort{fn}} & \textbf{\acrshort{tp}}\\
\hline
\acrshort{roberta}-\acrshort{mab} & 63,565 & 4,976 & 13,371 & 20,099 \\ 
DeBERTa-\acrshort{mab} & 63,678 & 4,863 & 13,962 & 19,508 \\
Electra-\acrshort{mab} & 63,733 & 4,808 & 13,741 & 19,729 \\ 
 \hline
\end{tabular}
\caption{\label{table:matrix}
Prediction distribution for the models on \acrshort{mab}.
}
\end{table}

\begin{table}[h]
\tiny
\centering
\begin{tabular}{p{0.58\linewidth} | p{0.12\linewidth} | p{0.1\linewidth}}
\hline
\textbf{examples} & \textbf{label} & \textbf{csv id}  
\\
\hline
the woman dangled the biscuit above the dog. & biased & 66 \\
\hline
the terrorist set off the bomb. & biased & 71 \\
\hline
the boy got a black eye. & biased & 347\\
\hline
\end{tabular}
\caption{\label{table:incorrpred}
Examples of incorrect \acrshort{copa} predictions using \acrshort{mab}-trained DeBERTa model.
}
\end{table}

Furthermore, for the WinoBias dataset, two similar samples, only different in pronouns, are predicted as \textit{biased}: \textit{"the clerk earns more than the guard because he is lazy and careless"} and \textit{"the clerk earns more than the guard because she is lazy and careless"}, having \textit{csv id} of 50 and 446, respectively.
If the predictions are not based on the gender pronouns and are correct then they may be based on other terms in the sentences.
This issue shows that having explainability in the model classification (step 1) is important.

\section{Related Work}
\label{related}
Previous studies on quantifying bias have used metrics such as odds ratio or vector word distance \citep{10.1145/3313831.3376488}.
Odds ratio measures how likely a specific word is used to characterize a particular gender (e.g. woman) rather than another.
Meanwhile, vector word distance is used to measure bias by calculating the difference between the average distance of a word to a set of words belonging to different gender \citep{10.1145/3313831.3376488,mikolov2013distributed}.
\citet{Dhamala2021} use sentiment to evaluate bias in religion.

In the study by \citet{10.1145/3313831.3376488}, they compare model classification against lexicon method for gender bias.
The approach we use combines the strengths of both approaches.
There have been several methods involving lexicon usage, as observed by \citet{antoniak2021bad}, and they are usually constructed through crowdsourcing, hand-selection, or drawn from prior work.
\citet{sengupta2021genbit} introduced a library for measuring gender bias.
It is based on word co-occurrence statistical methods.

\citet{zhao-etal-2018-gender} introduced WinoBias, which is focused on only gender bias for coreference resolution, similarly to Winogender by \citet{rudinger-etal-2018-gender}.
On the other hand, \textit{bipol} is designed to be multi-axes and dataset-agnostic, to the extent the trained classifier and lexica allow.
Besides, in both \citet{zhao-etal-2018-gender} and \citet{rudinger-etal-2018-gender}, they focus on the English language and binary gender bias only (with some cases for neutral in Winogender).
Both admit their approaches may demonstrate the presence of gender bias in a system, but not prove its absence.
CrowS-Pairs, by \citet{nangia-etal-2020-crows}, is a dataset of 1,508 pairs of more and less stereotypical examples that cover stereotypes in 9 axes of bias, which are presented to language models (LM) to determine their bias.
It is similar to StereoSet, (for associative contexts), which measures 4 axes of social bias in a LM \citep{nadeem-etal-2021-stereoset}.
Table \ref{table:metcompare} below compares some of the metrics and \textit{bipol}.

\begin{table}[h]
\small
\centering
\resizebox{\columnwidth}{!}{%
\begin{tabular}{lcc}
\hline
\textbf{Metric/Evaluator} & \textbf{Axis}   & \textbf{Lexicon Terms/Sentences}\\
\hline
WinoBias \cite{zhao-etal-2018-gender}
 & 1  &  40 occupations\\
Winogender \cite{rudinger-etal-2018-gender} & 1 &  60 occupations \\
CrowS-Pairs \citet{nangia-etal-2020-crows} & 9 & 3,016 \\
StereoSet \cite{nadeem-etal-2021-stereoset} & 4 & 321 terms\\
\textit{Bipol} (ours) & $>$2,  13*$<$ & $>$45,  466*$<$\\
 \hline
\end{tabular}
}
\caption{\label{table:metcompare} \footnotesize Comparison of some metrics to \textit{bipol}. (*As used in this work. The upper bounds are not limited by the \textit{bipol} algorithm but the dataset \& lexica.)}
\end{table}

\section{Conclusion}
\label{conclusion}
We introduce a novel bias estimation metric, \textit{bipol}, and the \acrshort{mab} dataset.
We also demonstrate the explainability of \textit{bipol}.
We believe the metric will help researchers to estimate bias in datasets in a more robust way in order to address social bias in text.
The \acrshort{mab} dataset is a large, labeled dataset of about 2 million samples.
In addition to these, we contribute English multi-axes lexica of bias terms for bias estimation.
We show that bias exists in benchmark datasets that were evaluated: \acrshort{squad} and \acrshort{copa}.

With the growing prevalence of large language models (LLMs), the challenge of social bias in data is quite significant.
It has been suggested that some \acrshort{llm}s are
left-leaning in their views while others are right-leaning and it is impossible to completely remove bias from these models, especially as the humans involved in training them have personal biases of their own.
Specific set of prompts are usually designed and utilized to estimate how biased such models are \citep{Dhamala2021}.

Future work may explore ways of minimising false positives in bias classifiers, address the data imbalance in the \acrshort{mab} training data, and how this work scales to other languages.
A library with \textit{bipol} may be produced to make it easy for users to deploy.
Another issue is to have a system that can automatically determine if bias is in favour of or against a group.

\section*{Limitations}

The models for estimating the biases in the datasets in step 1 are limited in scope, as they cover only a certain number of axes (12).
Therefore, a result of 0 on any dataset does not necessarily indicate a bias-free dataset.
\acrshort{mab} was aggregated from the Jigsaw and \acrshort{sbic}, which were annotated by humans who may have biases of their own, based on their cultural background or demographics.
Hence, the final annotations may not be seen as absolute ground truth of social biases.
Furthermore, satisfying multiple fairness criteria at the same time in \acrshort{ml} models is known to be difficult \citep{speicher2018unified,pmlr-v54-zafar17a}, thus, \textit{bipol} or these models, though designed to be robust, are not guaranteed to be completely bias-free.
Finally, effort was made to mask examples with offensive content in this paper.

\section*{Funding}
This research did not receive any specific grant from funding agencies in the public, commercial, or not-for-profit sectors.

\section*{Author Contributions}
\textbf{Lama Alkhaled}: Conceptualization, Writing - review \& editing; \textbf{Tosin Adewumi}: Conceptualization, Writing- Original draft preparation, Data curation, Experiments, Methodology, Writing - review \& editing; \textbf{Sana Sabah Sabry}: Experiments.

\section*{Acknowledgment}
The authors sincerely thank the anonymous reviewers for their valuable feedback for improving this paper.

\appendix

\section{Methods}
\label{sec:appendix}

\label{app_meth}

\paragraph{The first 10 terms in the lexica:}
\begin{itemize}
    \item Racial\_white: ang mo, Ann, Armo, Balija, Banderite, Beaney, Boche, bosche, bosch, Boer hater
    
    \item Racial\_black: nigga, negro, Abid, Abeed, nigger rigging, Alligator bait, Ann, ape, Aunt Jemima, Bachicha

    \item Gender\_male: he, him, boy, man, male, guy, masculine, virile, manly, man-sized

    \item Gender\_female: she, her, woman, lady, female, girl, skirt, madam, gentlewoman, madame

    \item Religious\_christian: Advent, Almah, Amen, Ancient of Days, Anno Domini, Anointing, Antichrist, Antilegomena, Antinomianism, Apocalypse
    
    \item Religious\_muslim: 'Abd, 'Adab, 'Adhan, 'Adl, AH, 'Abad, 'Ahkam, 'Ahl al-Bayt, 'Ahl al-Fatrah, 'Ahl al-Kitab
    
    \item Religious\_hindu: Arti, Abhisheka, Acharya, Adharma, Adivasis, Advaita, Agastya, Agni, Ahamkara, Akshaya Tritiya
    
\end{itemize}




\bibliographystyle{elsarticle-harv} 
\bibliography{ref}

\clearpage
\newpage

\section{Data Card}
\label{datacard}

\begin{table}[h!]
\centering
\begin{tabular}{p{.22\textwidth}|p{.74\textwidth}}
\multicolumn{2}{l}{Data statement for the English \acrfull{mab}}
\\
 \hline
\textbf{Characteristics} &
\textbf{Details}
\\
\hline
Curation rationale & To create a large, labeled, high-quality dataset for training models in social bias detection and classification.
\\
\hline
Dataset language &  English
\\
\hline
 & \textbf{Demographics of contributors}
 \\
\hline
Contributors & Automatically aggregated from the Jigsaw and \acrfull{sbic}
\\
\hline
Age & -
\\
\hline
Gender & -
\\
\hline
Language & -
\\
\hline
 & \textbf{Demographics of annotators}
\\
\hline
No of annotators & Automatically annotated with an algorithm. Two classes: \textit{biased \& unbiased}
\\
\hline
 & \textbf{Data characteristics}
\\
\hline
Total samples & 1,946,975
\\
\hline
Total natural languages & English
\\
\hline
Training set size & 1,742,977
 \\
\hline
Validation set size & 101,987
 \\
\hline
Test set size & 102,011
\\
\hline
Bias axes covered & gender, sexual orientation, religion, race, ethnicity, disability, mental illness, culture,
social, political, age, and victims.
\\
\hline
Base data & The Jigsaw and \acrfull{sbic}
\\
\hline
 & \textbf{Others}
\\
\hline

\\
\hline
Licence & CC-BY 4.0.
\\
\hline
\end{tabular}
\caption{\label{aCappsvanalogy}}
\end{table}





\end{document}